\documentclass[letterpaper, 10 pt, conference]{ieeeconf}  % Comment this line out
\IEEEoverridecommandlockouts                              % This command is only
\usepackage[utf8]{inputenc}
\usepackage[T1]{fontenc}
\usepackage{graphicx} 
\usepackage{url}

\title{\LARGE \bf
Autonomous Driving Data Chain \& Interfaces
}

\author{Benjamin Kahl% <-this % stops a space
\thanks{This work was not supported by any organization}% <-this % stops a space
}

\begin{document}

\maketitle
\thispagestyle{empty}
\pagestyle{empty}

%%%%%%%%%%%%%%%%%%%%%%%%%%%%%%%%%%%%%%%%%%%%%%%%%%%%%%%%%%%%%%%%%%%%%%%%%%%%%%%%
\begin{abstract}

Recent developments in autonomous driving technology have proven that map data may be used, not only for general routing purposes, but also for to enhance and complement common sensor data. This document reviews the most commonly used interfaces and formats at each step of a self-healing map data chain.

\end{abstract}

%%%%%%%%%%%%%%%%%%%%%%%%%%%%%%%%%%%%%%%%%%%%%%%%%%%%%%%%%%%%%%%%%%%%%%%%%%%%%%%%
\section{Introduction}

The utilization of digital maps on smartphones and navigation devices has become a widely common practice in many areas of the world. Digital mapping services, such as those offered by {\it{Here WeGo}}, {\it{TomTom}} or {\it{Google}}, have made navigational instructions based on simple road descriptions accessible to a large consumer basis.

However, the modern surge of interest in advanced driver assistance systems and autonomous vehicles has brought forth an ever increasing need for a whole new set of maps, purposefully built for robotic systems. 

The commonly termed {\it{High Definition Maps}}, or simply {\it{HD Maps}}, are advanced road descriptions specifically built for self-driving purposes. Including highly valuable information down to centimeter precision such as lane- and road-boundaries, these maps can provide indispensable data to be incorporated into the wider sensor-fusion concept.

\section{Autonomous Driving Setups}

\subsection{Static map caching}

Self-driving vehicles are typically equipped with a plethora of sensors aiding them in building a virtual model of their surrounding environment. Commonly employed sensors such as LiDARs, cameras and radars feed their gathered measurements into an on-board, intelligent controller program which identifies any objects around the vehicle, determines the correspondingly desired driving action and sets the cars actuators accordingly.

The limitations of this rudimentary setup emerge from the intrinsic restrictions common sensors are bound to:

\begin{itemize}
    \item Most consumer-grade LiDAR scanners can only provide reliable data in a range of up to 300 meters. \cite{lidrad}
    \item Due to sub-pixel disparity, the metric distance error in stereoscopic cameras increase non-linearly in relation of object distance. \cite{ster1} \cite{ster2}
    \item Radar sensors cannot provide reliable measurements on an objects size and suffer from similar range limitations as LiDAR scanners. \cite{lidrad}
\end{itemize}{}

Furthermore, none of the above listed sensors are capable of measuring data through solid obstacles such as other vehicles, trees or buildings.

To remedy these limitations a map can be employed, providing data well beyond the immediate vicinity of the vehicle. Indeed, maps are becoming increasingly popular in many automated driving systems and are often integrated as an additional sensor, complementing the data measured by conventional means.

\subsection{Hybrid map streaming}

As is the case with HD maps, a vast number of inaccuracies can arise from a single, static copy when the environment is subject to changes.

Thus, this system may be evolved to employ a real-time streaming approach in which current map data is continuously transmitted to the car from a dynamic content repository.
As an extension, a locally cached copy of the most recent version can be kept on-board the vehicle at all times, with updates being fetched in regular intervals.

This {\it{hybrid}}, cloud-based approach entails a permanent, low-latency and high bandwidth connection, providing a significant advantage over static methods only if host maps are regularly updated.

\subsection{Self-healing map stream}

\begin{figure}[h]
    \centering
    \includegraphics[width=.5\textwidth]{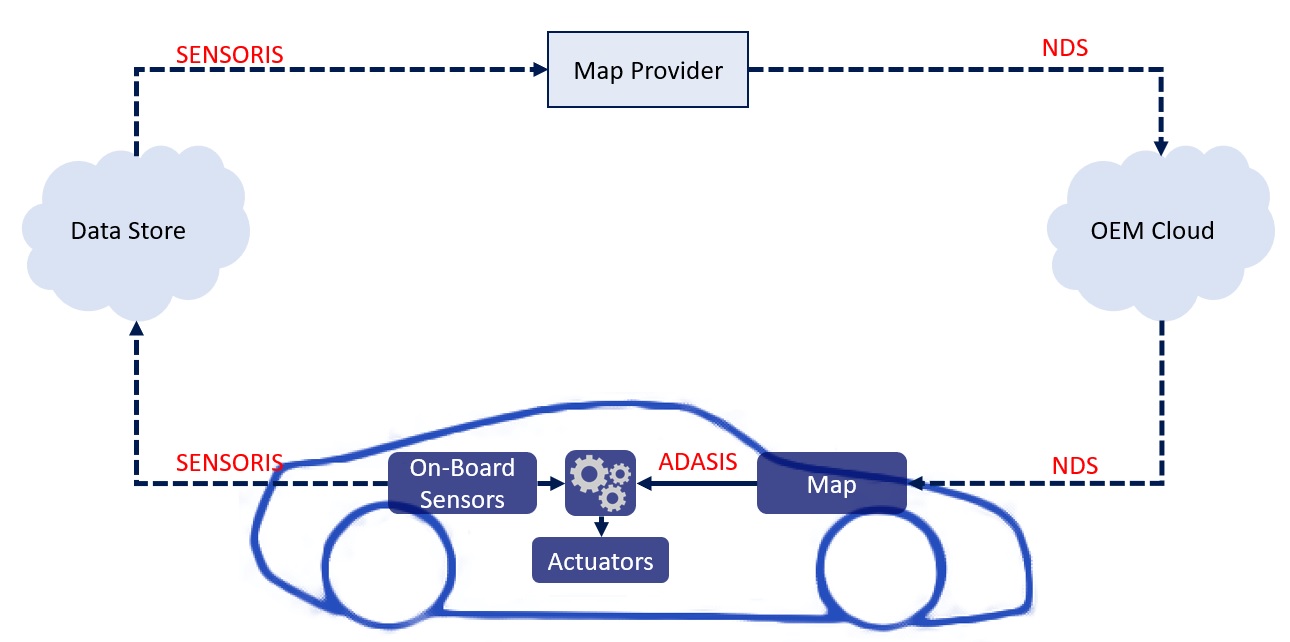}
    \caption{Self-healing map data chain and most commonly used interfaces.}
    \label{fig:gen_setup}
\end{figure}

Steady corrections and updates to a provided map database can be achieved through sensor-crowdsourcing. Here, the sensor data collected by vehicles is sent back to the map provider and used to repair and renew outdated information. This type of data chain is commonly called a {\it{self-healing}} setup and is depicted in fig. \ref{fig:gen_setup}.

The subsequent sections will focus on the individual steps of generalized map production as well as the cloud-to-vehicle and vehicle-to-cloud interfaces.

\section{Map Production}

The production of high-accuracy maps is a complex process that begins with the collection of vast amounts of raw data.

Specially designed mapping vehicles that are equipped with high-quality LiDAR, precision GPS and further critical components can garner a large sum of detailed information about the environment. Combining this with satellite imagery, government provided data and legacy maps provides a solid basis for the desired HD map.

After going through a selection and verification process, sensor-crowdsourced data is organized, harmonized and finally merged into the published map layers. Sensors that may provide useful data range from LiDAR and image data to V2X sensors.

\begin{figure}[h]
    \centering
    \includegraphics[width=.45\textwidth]{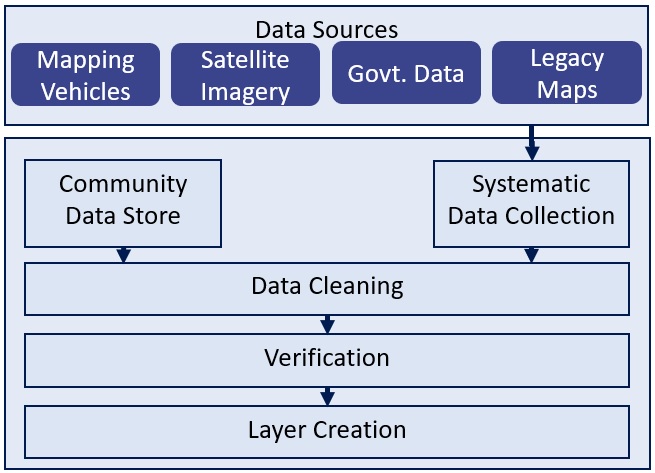}
    \caption{Common steps in the production of HD maps.}
    \label{fig:map_prod}
\end{figure}

The result is a complete map-database containing a global model of all collected information. The de-facto standard used for databases of these types is the {\it{Navigation Data Standard}} (NDS).

\section{Navigation Data Standard}

The Navigation Data Standard is a standardized format for navigation databases which aims to enable a seamless fusion of map and sensor data for advanced driving automation. It is being developed and maintained by the NDS association consisting of {\it{BMW}}, {\it{Volvo}}, {\it{TomTom}} and several more members.

In order to enable its usage in autonomous driving systems, NDS maps strive towards providing following key benefits over their peers:

\begin{itemize}
    \item Independence from vehicle or navigation-system manufacturer.
    \item Flexible content management and consistent versioning.
    \item Allow incremental updates to be delivered to select regions.
    \item Serve a variety of use cases.
\end{itemize}

In order to accommodate these requirements, NDS uses the {\it{SQLite}} database file format as an underlying basis. This enables NDS databases from different suppliers to be merged and consistently versioned.

Each product database covers a specific geographic area (Example: {\it{Europe basic navigation}} by TomTom) which is divided into a multitude of update regions. Update regions (Example: Germany) can be subject to incremental or partial updates. Similarly to regular version-control repositories, each map update is delivered in the form of change-only information, which dramatically reduces the size of content updates and thus enables over-the-air map transmissions.

All data ultimately belongs to a specific building block or layer, each of which addresses a specific domain of data. Typical building blocks would be: Traffic information, Routing data, Points of interest, etc.

Each individual application would filter the required building blocks from the aggregate database in order to cover the intended use case. Of particular importance to automated driving are the building blocks related to lane geometry, volatile data and 3D geometry.

\begin{figure}[h]
    \centering
    \includegraphics[width=.5\textwidth]{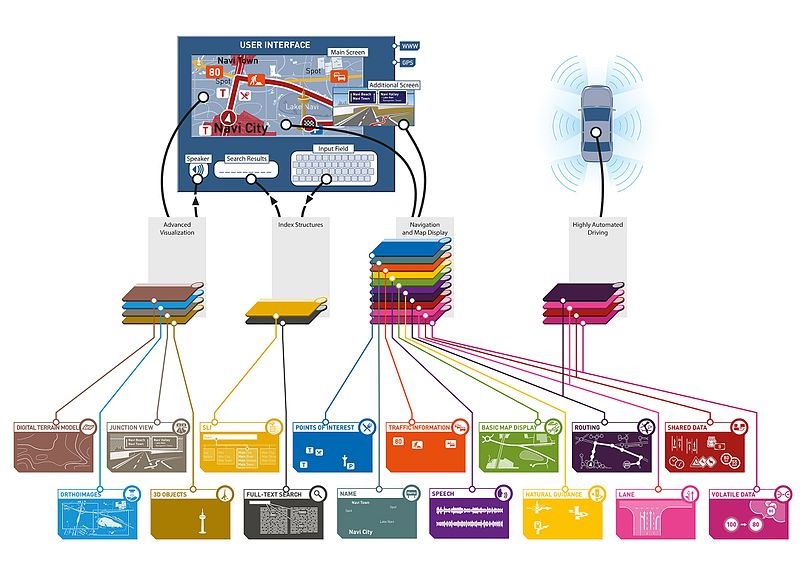}
    \caption{NDS building blocks as portrayed by {\it{Emde Grafik}}.}
    \label{fig:nds}
\end{figure}

These types of large databases are often clumsy and costly to handle by real-time applications. For instance, the exact nature of speed signage on the other side of a country are of little concern to an autonomous driving program. 

The next section describes a standardized interface meant to provide autonomous driving applications only with the relevant vicinity of the vehicle.

\section{ADASIS Interface}

Advanced in-vehicle applications such as headlight control, infotainment software or even fully autonomous driving are all commonly placed under the {\it{ADAS}} umbrella (Advanced driver assistance systems). 

The data requirements of these systems are often limited to a vehicles surrounding environment, known as the {\it{ADAS Horizon}}. Given the wide range of use cases, the {\it{ADASIS Forum}} was launched in order to establish an industry wide standard for the provision of such data.

\subsection{ADASIS Forum}

The ADASIS Forum was formed in May 2001 with the primary goal of developing a standardized map data interface between stored map data (NDS databases) and rudimentary ADAS applications.

The forum is hosted and coordinated by {\it{ERITCO}}, including prominent members from the automotive industry such as {\it{BMW}}, {\it{Ford}} and {\it{Opel}}.

After its release in 2010, the ADASIS v2 specification has become the widely accepted de-facto standard for ADAS Horizon Providers, with 2011 giving light to the first products making use of this interface. 

A third version of the specification was launched in 2018, specifically targeted at autonomous driving systems.

\subsection{ADAS Horizon}

The immediate vicinity of the vehicle that is extracted from an NDS database and passed on to individual ADAS applications is known as the ADAS Horizon.

The extent of the calculated horizon is derived from {\it{the road ahead}}, which equates to the {\it{Most Probable Path}} (MPP) the car is likely to take. Each road segment is assigned a probability of the car driving through it, depending on each individual use case.

\begin{figure}[h]
    \centering
    \includegraphics[width=.5\textwidth]{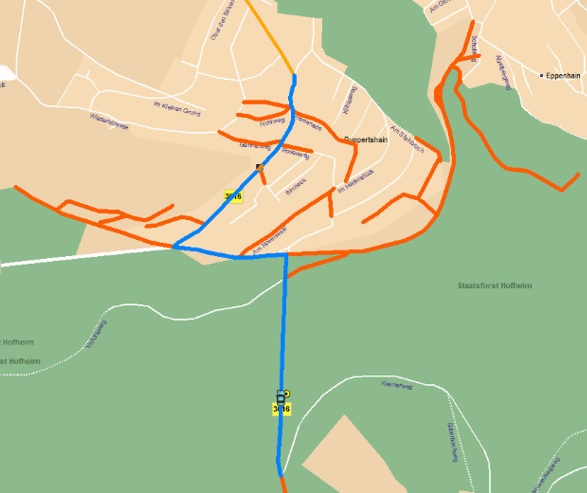}
    \caption{MPP (in blue) and ADAS horizon (in orange) as shown by Ress et al.\cite{main}}
    \label{fig:nds}
\end{figure}

\subsection{ADASIS System Architecture}

The ADAS horizon is extracted from an NDS map database by a so-called {\it{ADAS Horizon Provider}} and subsequently delivered on a CAN (Control Area Network) bus system. The horizon mainly consists of road segments and their respective attributes. The receiving applications decode the message with a {\it{Horizon Reconstructor}}.

\begin{figure}[h]
    \centering
    \includegraphics[width=.45\textwidth]{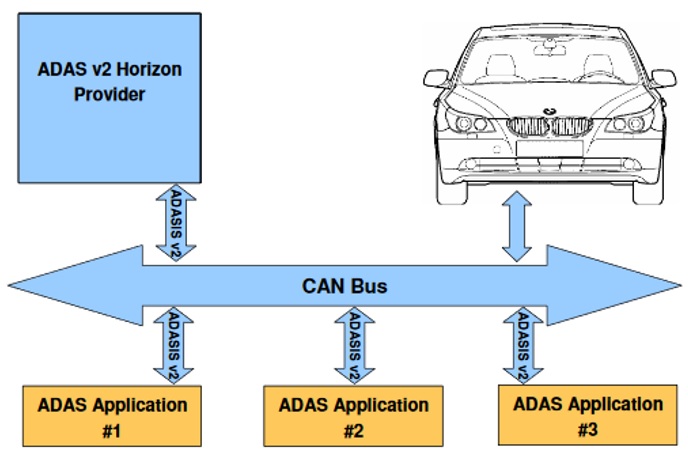}
    \caption{Core components of the ADASIS architecture as portrayed by Ress et al.\cite{main}}
    \label{fig:arch}
\end{figure}

Thus, the core entities of an ADAS System could be condensed to the following few component types:

\begin{itemize}
    \item ADAS Horizon Provider (AHP), which extracts the ADAS horizon and sends it to the respective applications.
    \item The ADASIS Protocol, which describes how the horizon is to be encoded or decoded at each end of the communication line.
    \item An on-board vehicle CAN bus which serves as a communication interface for each component.
    \item ADAS applications, each with their own Horizon Reconstructor (AHR).
\end{itemize}{}

A system of this architecture scales poorly with the amount of data being transmitted, particularly if the underlying CAN bus supports no multiplexing.

As a result, the primary focus in the ADASIS v2 specification is the minimization of bus load.

\subsection{ADASIS v1 and v2}

The ADASIS v1 specification was implemented as a prototype and proof of concept. It having successfully passed performance and interoperability tests, the Forum moved on to the development of the v2 standard intended for implementation by an OEM.

\begin{figure}[h]
    \centering
    \includegraphics[width=.5\textwidth]{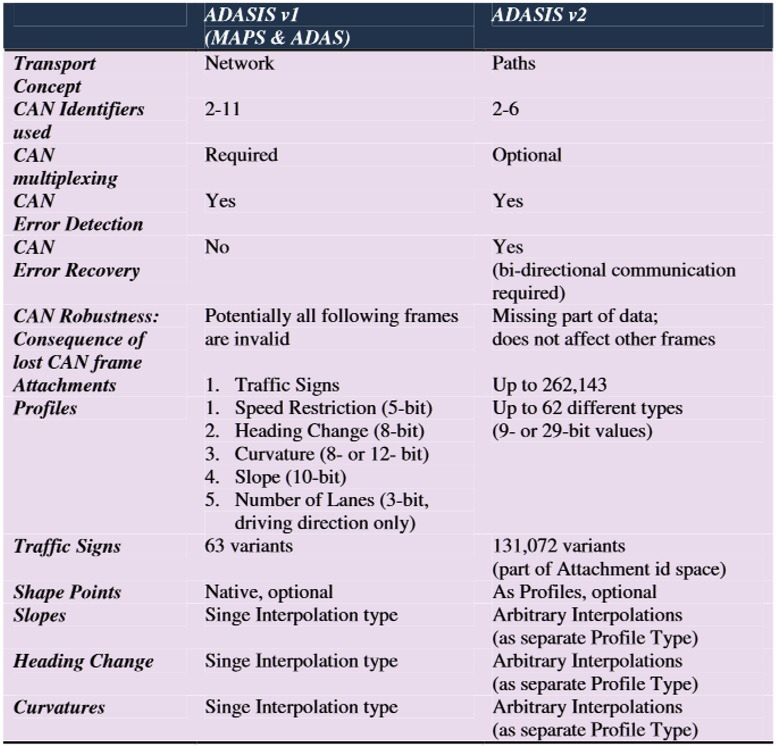}
    \caption{Core differences between the ADASIS v1 and v2 specifications, as documented by Ress et al.\cite{main}}
    \label{fig:v1v2}
\end{figure}

Fig. \ref{fig:v1v2} lists some of the primary differences between the two specifications. Most of these advances concern the flexibility of the CAN bus communication-line, with v2 not requiring a polymorphic CAN frame but enabling the recovery of corrupted data upon CRC error detection should multiplexing be available.

Further improvements, such as differentiation between an increased number of traffic sings, attachment profiles and curvature interpolations are also available.

The most important difference, however, lies in the utilized data structure for the representation of the horizons roads (Transport Concept). The v1 specification maintains the horizon as a network, meaning a set of interconnected nodes where other paths may branch off from the position of a given node.

Instead, the main logical entity of the v2 specification is a simple {\it{PATH}} element, where other branching paths are treated just as any other attribute (for example traffic sings).

This concept greatly decreases the overall size of the transmitted data and is particularly adequate for horizons based on a single path concept.

\subsection{ADASIS v2 Message types}

The ADAS horizon is provided in the form of a message stream, where the first three bits of each message are reserved for CRC purposes in order for the Horizon Reconstructors to detect missing data. If a bidirectional communication is available, Reconstructors may issue a re-transmission request to the Horizon Provider, ensuring the robustness of the system.

\subsubsection{Segment Message}

An ADAS horizon consists primarily of a series of segment objects, where each path is made up of one or more non overlapping segments. Each segment includes a reference index of its respective path as well as a plethora of information about the underlying road.

Effective speed limits, number of lanes as well as the route type are all included, in addition to a series of boolean fields designating the segment to be part of a tunnel, bridge or emergency lane.

\subsubsection{Profile Message}

A profile object determines the exact curvature of a segment, where the {\it{Value0}} field indicates the profile value at {\it{Offset}} and the {\it{Value1}} field indicates the profile value at {\it{Offset+Distance1}}, where {\it{Distance1}} equates to the pre-calculated length of the curve.

Profiles can be discrete (zero order), linear, quadratic or of a higher order with certain interpolation types (such as bezier-curves) requiring additional control points.

\subsubsection{Attachment Message}

Attachment objects represent signs or similar attributes on a path. The path is referenced via an index with an interpolation number (or offset) designating the exact location on said path. The {\it{Offset}} variable is stored as a 13 bit number, limiting the given accuracy to $\frac{1}{13^2}*(path length)$.

\begin{figure}[h]
    \centering
    \includegraphics[width=.45\textwidth]{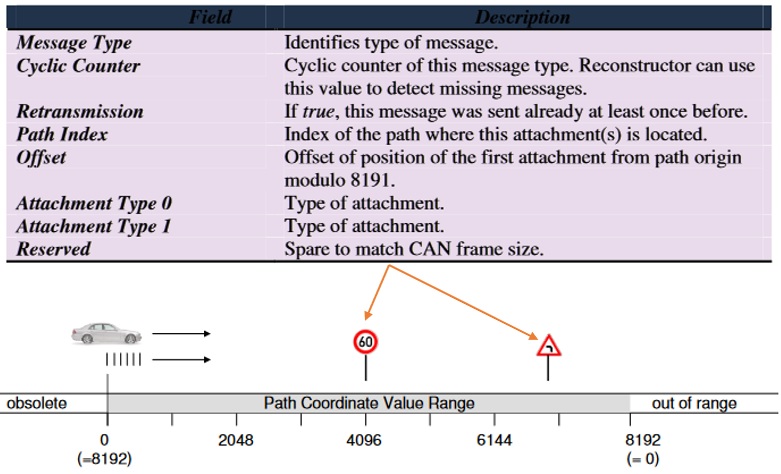}
    \caption{Composition of an attachment message, as documented by Ress et al.\cite{main}}
    \label{fig:att}
\end{figure}

\subsubsection{Stub Message}

As stated earlier, the ADASIS v1 specification requires a separate node for each path junction, while v2 allows the usage of {\it{STUB}} objects. Stubs act in almost the exact same way as regular attachments by referencing a certain point on a path through an index and offset variable. However, instead of a single attribute like a traffic sign, stubs point to a path that branches off at the given location.

Furthermore, the angular predicament, number of lanes and MPP probability are also referenced. 

This results in a far simpler reconstruction process than would be the case in a network dataset. Whether if the paths referenced by the given stubs are also transmitted depends on whether if the OEM prefers a single path or multi path horizon.

\subsubsection{Position Message}

In a similar fashion, the vehicles current position is also stored through a path index and offset value with a position probability and confidence value indicating the certainty of the given information. In addition, a GPS timestamp, vehicle speed value and the current lane are also included in a position message.

\subsection{ADASIS v3}

The ADASIS v3 specification was released to all ADASIS Forum members in 2018 and is specialized for use with autonomous driving applications, incorporating native support for HAD maps and significantly larger horizons. Detailed information such as precise lane boundaries, 3D obstacle descriptions and geometry lines for guardrails are also being extracted from NDS databases and transmitted to ADAS applications.

Finally, the v3 specification allows local changes to be made on cached map data such that faulty information can be modified on the fly by the control program.

\section{Sensoris Interface}

The {\it{Sensor Interface Specification}} (SENSORIS) is the preferred standard for sensor-crowdsourcing purposes. The specification differentiates between three layers of actors: Individual vehicles belong to a {\it{vehicle fleet}} which communicates with a corresponding {\it{vehicle cloud}}. Multiple vehicle coulds communicate with a single {\it{service cloud}}.

Common use cases of this interface include but are not restricted to:

\begin{itemize}
    \item Real-time traffic, weather etc.
    \item Map repair and healing.
    \item Statistical data analysis.
\end{itemize}{}

\subsection{Sensoris Message Types}

Sensoris defines three types of messages transmitted between vehicle fleets and clouds:

\begin{itemize}
    \item {\it{Data messages}} contain vehicle sensor data of one of three hierarchical definition-classes. {\it{Standard Definition}} (SD) for GPS, odometry and gyro data, {\it{High Definition}} (HD) for video, radar and ultrasonic data, {\it{Automated Driving}} (AD) for 360-degree video and LiDAR data.
    \item {\it{Job request messages}} can be described as data requests by OEMs made to the vehicles. The job requests contain the requested type of sensor as well as a detailed set of constraints.
    \item {\it{Job  status  messages}} contain the current status of job requests.
\end{itemize}{}

\begin{figure}[h]
    \centering
    \includegraphics[width=.45\textwidth]{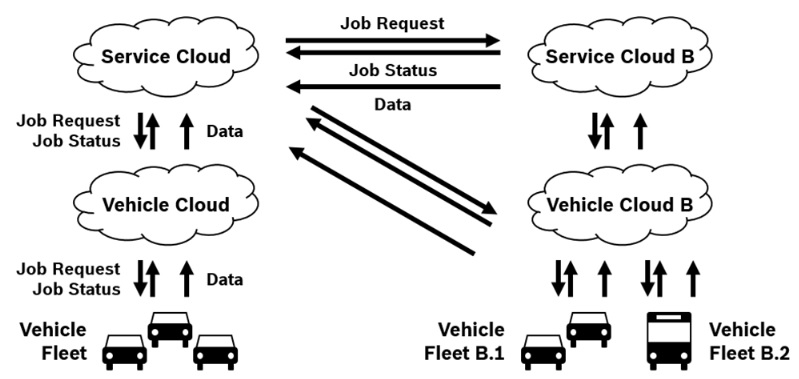}
    \caption{Multiple actor roles and interface as seen in the Sensoris v1.0 specification \cite{sens}}
    \label{fig:sensoris}
\end{figure}

\subsection{Sensoris Architecture}

The interface architecture limits itself to defining only content and encoding methods in order not to restrict different implementation methods and be agnostic to the communication channel used.

There is an emphasis on the minimization of serialized data by choosing a compact data format. The Sensoris v1.0 specification recommends either {\it{Apache  Avro}}, {\it{Apache  Thrift}} or {\it{Google Protocol Buffers}}, as these fulfill all necessary requirements.

\section{CONCLUSIONS}

The NDS format builds on modular SQLite data structures, enabling dynamic feature upgrades, such as the download of specific content layers while keeping the required connection bandwidth reasonably low by utilizing change-only updates. NDS maps work worldwide with group members offering coverage over most geographic regions. The format has proven itself through a variety of consumer products available since 2012, including BMW and Volkswagen cars.

On the other hand, ADASIS and SENSORIS, both part of the Open AutoDrive Forum, are being supported on an industry-wide basis by a multitude of OEMs, but only few implementation have actually been deployed. The problematic bus load of ADASIS v1 was addressed in its second iteration but concerns about costly bus load due to the increased horizon and data accuracy in ADASIS v3 remain largely unaddressed. SENSORIS provides a flexible, yet clearly defined specification but also lacks completed and deployed products making use of it.

Further testing may be required before passing a final verdict on the suitability of these interfaces in self-healing autonomous driving setups.

\addtolength{\textheight}{-12cm}   % This command serves to balance the column lengths

%%%%%%%%%%%%%%%%%%%%%%%%%%%%%%%%%%%%%%%%%%%%%%%%%%%%%%%%%%%%%%%%%%%%%%%%%%%%%%%%

%%%%%%%%%%%%%%%%%%%%%%%%%%%%%%%%%%%%%%%%%%%%%%%%%%%%%%%%%%%%%%%%%%%%%%%%%%%%%%%%

%%%%%%%%%%%%%%%%%%%%%%%%%%%%%%%%%%%%%%%%%%%%%%%%%%%%%%%%%%%%%%%%%%%%%%%%%%%%%%%%

\end{document}